\documentclass[10pt,twocolumn,letterpaper]{article}

\usepackage[pagenumbers]{cvpr} 

\newcommand{\teaminfo}[3]{
    \begin{table}[H]
        \vspace{-2mm}
        \begin{tabularx}{\linewidth}{@{}lX@{}}
            \toprule
            \textbf{Title:} & {#1} \\
            \textbf{Members:} & {#2} \\
            \textbf{Affiliation:} & {#3} \\
            \bottomrule
        \end{tabularx}
        \vspace{-2mm}
    \end{table}
    }

\usepackage{multirow}
\usepackage{colortbl}
\usepackage{xcolor}
\usepackage{pifont}
\usepackage{makecell}
\usepackage{adjustbox}
\usepackage{threeparttable}
\usepackage{setspace}
\usepackage{tabularx}
\usepackage{float}

\usepackage{pythonhighlight}
\usepackage[linesnumbered,ruled,vlined]{algorithm2e}

\definecolor{cvprblue}{rgb}{0.21,0.49,0.74}
\usepackage[pagebackref,breaklinks,colorlinks,allcolors=cvprblue]{hyperref}

\def\paperID{*****} 
\usepackage[misc]{ifsym}
\def\confName{CVPR}
\def\confYear{2025}
\usepackage[accsupp]{axessibility}  
\usepackage{fontawesome5}
\usepackage{array}
\definecolor{gold}{RGB}{255, 205, 0}
\definecolor{silver}{RGB}{192, 192, 192}
\definecolor{bronze}{RGB}{205, 127, 50}  
\title{PVUW 2025 Challenge Report:\\ Advances in Pixel-level Understanding of Complex Videos in the Wild}

\author{Henghui Ding\textsuperscript{*}, Chang Liu\textsuperscript{*}, Nikhila Ravi\textsuperscript{*}, Shuting He\textsuperscript{*}, Yunchao Wei\textsuperscript{*}, Song Bai\textsuperscript{*}, Philip Torr\textsuperscript{*}\\
Kehuan Song,\quad Xinglin Xie,\quad Kexin Zhang,\quad Licheng Jiao,\quad Lingling Li,\quad Shuyuan Yang\\
Xuqiang Cao,\qquad Linnan Zhao,\qquad Jiaxuan Zhao,\qquad Fang Liu\\
Mengjiao Wang,\qquad Junpei Zhang,\qquad Xu Liu,\qquad Yuting Yang,\qquad Mengru Ma\\
Hao Fang,\qquad Runmin Cong,\qquad Xiankai Lu,\qquad Zhiyang Chen,\qquad Wei Zhang\\
Tianming Liang,\qquad Haichao Jiang,\qquad Wei-Shi Zheng,\qquad Jian-Fang Hu\\
Haobo Yuan,\qquad Xiangtai Li,\qquad Tao Zhang,\qquad Lu Qi,\qquad Ming-Hsuan Yang\\
\href{https://pvuw.github.io/}{https://pvuw.github.io/}
}

\begin{document}
\maketitle
\renewcommand{\thefootnote}{\fnsymbol{footnote}}
\footnotetext[1]{Authors are CVPR 2025 PVUW Challenge organizers. All others are challenge participants from the top-3 teams of MOSE and MeViS tracks.}
\footnotetext[0]{${\textrm{\Letter}}$ Corresponding to Henghui Ding (henghui.ding@gmail.com), the Institute of Big Data, Fudan University, Shanghai, China.}

\begin{abstract}
This report provides a comprehensive overview of the 4th Pixel-level Video Understanding in the Wild (PVUW) Challenge, held in conjunction with CVPR 2025. It summarizes the challenge outcomes, participating methodologies, and future research directions. The challenge features two tracks: MOSE, which focuses on complex scene video object segmentation, and MeViS, which targets motion-guided, language-based video segmentation. Both tracks introduce new, more challenging datasets designed to better reflect real-world scenarios. Through detailed evaluation and analysis, the challenge offers valuable insights into the current state-of-the-art and emerging trends in complex video segmentation. More information can be found on the workshop website: \url{https://pvuw.github.io/}.
\end{abstract}    
\section{Introduction}
\label{sec:intro}

Pixel-level understanding of dynamic and complex visual scenes remains a core yet unresolved problem in computer vision~\cite{MOSE,MeViS,li2024transformer,wu2024towards,ravi2024sam,hesham2025exploiting}. While traditional research has predominantly focused on semantic segmentation within static images~\cite{CCL,BFP,SVC}, such approaches fall short in capturing the temporal continuity of the real world. In contrast, video segmentation~\cite{MOSE,MeViS,sun2022coarse,SunTPAMI,ke2022video,ke2023mask} offers a more realistic framework, aligning better with applications that demand spatiotemporal reasoning—such as autonomous driving, aerial navigation, and mobile video editing. These use cases underscore a growing shift toward scene understanding methods that are not only spatially precise but also temporally coherent. To advance research in this direction, we introduce the Pixel-level Video Understanding in the Wild (PVUW) workshop, which emphasizes the challenges posed by unconstrained, real-world environments~\cite{ding2024pvuw}. PVUW seeks to narrow the gap between static and dynamic scene understanding, encouraging the development of robust algorithms that can generalize across diverse, time-varying visual conditions. Through this initiative, we aim to catalyze innovation toward deploying perception systems capable of reliable operation in the wild.

Recent advances in Large Language Models and multimodal LLMs have significantly reshaped computer vision~\cite{shuai2024survey}. Alongside, foundational models like SAM2~\cite{ravi2024sam} have leveraged large-scale data to achieve strong generalization. Notably, progress in tasks such as Video Object Segmentation (VOS) \cite{MOSE} and Referring Video Object Segmentation (RVOS) \cite{MeViS} highlights the field’s continued momentum toward more robust and unified vision systems.

Building on these developments, the goal of our workshop and challenge is to keep pace with cutting-edge research, offer a challenging, yet realistic benchmark to evaluate state-of-the-art models, and provide valuable insights into both the current trends and future directions of video understanding. Following past challenges, we aim to continuously provide challenging and diverse benchmarking data that are taken in real world, and in this year, we have added more latest data that are first time released.

\section{The PVUW 2025 Challenge}

This year, we center our challenge around two focused tracks: the MOSE Track, which benchmarks advanced VOS methods in complex and densely populated scenes; and the MeViS Track, which evaluates models on language-guided video segmentation, with a particular emphasis on motion-guided language expressions.

\subsection{Two Video Segmentation Tracks}

\indent\textbf{Track 1: MOSE Track}

\begin{table}[t]
    \renewcommand\arraystretch{1.1}
    \centering
    \setlength\tabcolsep{10pt}
    \caption{MOSE Track results and top 20 of the final rankings.}
    \vspace{-3mm}
    \footnotesize
    {\begin{tabular}{rlccc}
            \toprule
             Rank & Team  & {$\mathcal{J}$} & {$\mathcal{F}$} & {$\mathcal{J\&F}$} \\
             \midrule
             \textcolor{gold}{\faTrophy} 1 & BrainyBots & 83.59 & 90.92 & 87.26 \\
             \textcolor{silver}{\faMedal} 2 & DeepSegMa & 82.50 & 90.07 & 86.28 \\
             \textcolor{bronze}{\faMedal} 3 & JIO & 80.28 & 87.57 & 83.92 \\
             4  & SCU\_Leung & 79.93 & 87.33 & 83.63 \\
             5  & wulutuluman & 79.89 & 87.21 & 83.55 \\
             6  & mima & 79.80 & 87.21 & 83.51 \\
             7  & LK186******96 & 79.80 & 87.10 & 83.45 \\
             8  & STELATOS9 & 79.65 & 87.16 & 83.41 \\
             9  & MaxBitter & 79.64 & 87.10 & 83.37 \\
             10 & XiaomiYU7 & 79.47 & 86.92 & 83.20 \\
             11 & menghaoran & 79.59 & 86.79 & 83.19 \\
             12 & zjy05140514 & 79.46 & 86.85 & 83.15 \\
             13 & keeper & 79.48 & 86.83 & 83.15 \\
             14 & zhaojinhui & 79.44 & 86.83 & 83.14 \\
             15 & LuxeedR7 & 79.40 & 86.85 & 83.12 \\
             16 & HuaweiAITOM9 & 79.23 & 86.58 & 82.91 \\
             17 & YuLinLin & 79.15 & 86.55 & 82.85 \\
             18 & ccHub & 78.93 & 86.68 & 82.80 \\
             19 & ZhiMu & 78.79 & 86.59 & 82.69 \\
             20 & ppbb & 78.69 & 86.11 & 82.40 \\
            \bottomrule
        \end{tabular}}%
    \label{tab:results_mose}%
\end{table}%

\noindent\textbf{\textit{Complex Video Object Segmentation (MOSE)}}~\cite{MOSE} aims to track and segment objects in videos of complex environments. This track is based on the MOSE~\cite{MOSE} dataset, which is a new video object segmentation benchmark designed to study object tracking and segmentation in complex, real-world scenes. Unlike previous video segmentation datasets~\cite{Pont-Tuset_arXiv_2017,youtube_vos} that focus on salient and isolated objects, MOSE features crowded environments, frequent occlusions, and object disappearances. It consists of 2,149 video clips and 5,200 objects across 36 categories, with over 430,000 high-quality segmentation masks. MOSE challenges existing VOS models and highlights the performance gap in complex scenarios, encouraging further research into robust segmentation techniques. This year's testing set is a part of MOSE testing set, but with more challenging newly taken data added. The ground truths of all videos in the testing sets are confidential and has never been released before. 
This year, we have 81 teams registered to the MOSE track on the platform, and 43 teams of them submitted their results on the testing phase. Top results are shown in Table \ref{tab:results_mose}. The top three teams are imaplus, KirinCZW, and dumplings. The first place team achieved a $\mathcal{J}\&\mathcal{F}$ score of 87.26\% on the testing set.

\vspace{2mm}
\noindent\textbf{Track 2: MeViS Track}

\begin{table}[t]
    \renewcommand\arraystretch{1.1}
    \centering
    \setlength\tabcolsep{10pt}
    \caption{MeViS Track results and top 20 of the final rankings.}
    \vspace{-3mm}
    \footnotesize
    {\begin{tabular}{rlccc}
            \toprule
             Rank & Team  & {$\mathcal{J}$} & {$\mathcal{F}$} & {$\mathcal{J\&F}$} \\
             \midrule
             \textcolor{gold}{\faTrophy} 1 & MVP-Lab & 58.83 & 65.14 & 61.98 \\
             \textcolor{silver}{\faMedal} 2 & ReferDINO-iSEE & 56.79 & 64.07 & 60.43 \\
             \textcolor{bronze}{\faMedal} 3 & Sa2VA & 52.68 & 59.84 & 56.26 \\
             4 & Pengsong & 53.06 & 58.76 & 55.91 \\
             5 & ssam2s & 52.00 & 58.33 & 55.16 \\
             6 & strong\_kimchi & 51.78 & 58.27 & 55.02 \\
             7 & seilvik90 & 50.61 & 59.22 & 54.91 \\
             8 & yiweima\_xmu & 50.93 & 58.65 & 54.79 \\
             9 & maclab & 50.63 & 58.32 & 54.48 \\
             10 & xinming & 51.24 & 57.33 & 54.28 \\
             11 & zhangtao-whu & 51.22 & 57.19 & 54.21 \\
             12 & yiweima & 50.49 & 57.30 & 53.90 \\
             13 & TransVG321 & 50.10 & 57.30 & 53.70 \\
             14 & xmu-xiaoma666 & 49.86 & 56.92 & 53.39 \\
             15 & MYOLO & 49.80 & 56.97 & 53.38 \\
             16 & j\_kker101 & 50.02 & 56.55 & 53.29 \\
             17 & X-CLIP & 49.64 & 56.84 & 53.24 \\
             18 & tbao & 49.05 & 56.59 & 52.82 \\
             19 & LuQiLXX & 48.48 & 54.69 & 51.59 \\
             20 & mengyuan & 48.63 & 54.42 & 51.53 \\
            \bottomrule
        \end{tabular}}%
    \label{tab:results_mevis}%
\end{table}%

\noindent\textit{\textbf{Motion Expression guided Video Segmentation (MeViS)}} \cite{MeViS} focuses on segmenting objects in video based on a sentence describing the motion of the objects, which is based on the MeViS dataset. The MeViS dataset~\cite{MeViS} is a large-scale benchmark designed for motion-guided language-based video object segmentation. Unlike previous referring image segmentation or referring video segmentation works~\cite{GRES,VLTPAMI,ding2021vision,ISFP,wang2025hierarchical,wu2024towardstip,M3Att,Zhang_2021_ICCV,he2024segpoint,he2023grec,liu2024primitivenet,liu2024referring,he2024refmask3d,3DGRES} that focus on static object attributes, MeViS emphasizes motion-centric language expressions to identify and segment target objects in complex video scenes. It features a wide range of motion expressions paired with videos containing crowded and dynamic environments. Benchmarking results show that existing referring video object segmentation methods struggle with this task, highlighting the need for new methods that can better leverage motion as a primary cue in language-guided video segmentation. Similarly, the testing set of this track comes from MeViS testing set, with newly added videos and confidential ground-truths. For MeViS Track, this year we have attracted 77 teams to registered, from which 31 teams participated in the testing phase. The top three teams are MVP-Lab, ReferDINO-Plus, and HarborY, as shown in Table \ref{tab:results_mevis}.

\subsection{Evaluation}
Both tracks are evaluated using standard metrics consistent with prior PVUW challenges~\cite{ding2024pvuw,ding2024lsvos} and benchmarks such as DAVIS~\cite{Pont-Tuset_arXiv_2017} and YouTube-VOS~\cite{youtube_vos}. Specifically, we adopt region similarity ($\mathcal{J}$), contour accuracy ($\mathcal{F}$), and their average ($\mathcal{J}\&\mathcal{F}$), with $\mathcal{J}\&\mathcal{F}$ serving as the primary ranking metric. All evaluations are conducted on the publicly accessible CodaLab platform.

\cref{sec:mose_method} and \cref{sec:mevis_method} presents the solutions from the top-3 teams of MOSE track and MeViS track, respectively.
\section{MOSE Track Top Solution}
\label{sec:mose_method}


\subsection{1st Team in MOSE Track: BrainyBots}

\teaminfo{STSeg}{Kehuan Song, Xinglin Xie, Kexin Zhang, Licheng Jiao, Lingling Li, Shuyuan Yang}{Xidian University, China}

We optimize our solution across both training and inference stages. During training, we fine-tune SAM2 and TMO on the MOSE dataset to better adapt them to the challenges of video object segmentation in complex environments. For inference, we leverage an ensemble of five models—SAM2, TMO, Cutie, XMem, and LiVOS—on the MOSE test set. The predicted masks from these models are aggregated to construct rich pseudo-labels. Based on these, we dynamically select the most suitable model per video instance to ensure optimal segmentation quality. Detailed fine-tuning strategies are provided in our full report.

\begin{figure}[t]
    \centering
    \includegraphics[width=\linewidth]{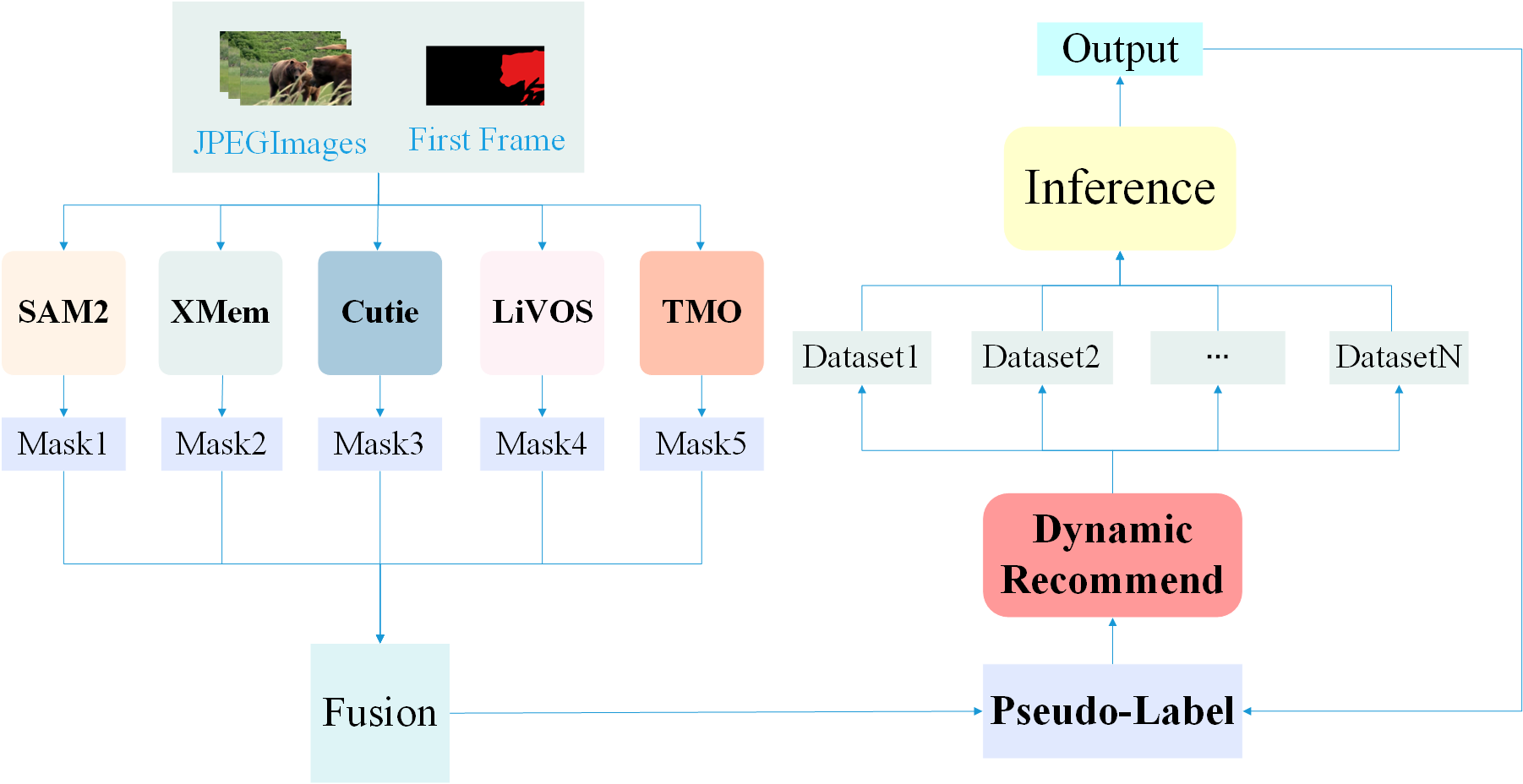} 
    \vspace{-3mm}
    \caption{Overview of the PGMR Framework. Inference and Pseudo-Label-Based Model Selection: Employing five models to conduct inference operations, and the model with optimal performance for different video contents is intelligently selected.}
    \label{fig:PGMR_framework} 
\end{figure}

\textbf{Adaptive Pseudo-labels Guided Model Refinement Pipeline}
After analyzing the dataset, we found it challenging to achieve good results in all scenarios using a single model. Therefore, we propose an Adaptive Pseudo-labels Guided Model Refinement Pipeline (PGMR), as shown in Fig \ref{fig:PGMR_framework} with specific implementation steps as follows:

\textbf{Multi-Model Inference: Independent Processing and Result Collection.}
In video frame segmentation and tracking tasks, we first employ multi-model independent inference to process the same set of video frames. Each model demonstrates unique performance advantages in different scenarios based on its design features and training data. To fully leverage the strengths of each model, we have designed a parallel inference framework that ensures each model can operate independently and produce results without interference from other models. This framework allows multiple models to perform inferences on the same set of video frames simultaneously, enabling each model to perform at its best without being influenced by others.
The output results of each model are collected separately and include segmentation masks, tracking IDs, and confidence scores. Segmentation masks are used to accurately delineate the boundaries of target objects within video frames while tracking IDs are employed to continuously track the positional changes of target objects throughout the video sequence and confidence scores reflect the model's assessment of each prediction. 

\textbf{Pseudo-Label Fusion: Generating a Baseline Result.}
To optimize the performance of video frame segmentation and tracking tasks, it is crucial to integrate the inference results of multiple models into a comprehensive pseudo-label. This pseudo-label serves as a key baseline for the subsequent optimization process and helps identify the model that performs optimally for different video contents. The generation of the pseudo-label involves several steps:

\begin{itemize}
    \item Firstly, a consistency check is carried out by comparing the segmentation masks and tracking IDs of different models to identify the regions where the model results are consistent and those where they are inconsistent.
    \item Then, confidence weighting is performed. Weights are assigned to each model based on its historical performance and the confidence scores associated with its predictions. 
    \item Finally, a voting mechanism is employed for the regions where the models produce conflicting results, and a conflict resolution strategy is adopted. 
\end{itemize}

The fused pseudo-label, as a key intermediate link, bridges the gap between the outputs of individual models and the performance of the unified optimization system.
It enables the intelligent selection of the model that demonstrates the best performance for different video contents. 

\textbf{Model Recommendation Mechanism: Intelligent Task Allocation.}
Based on the generated pseudo-label, we have developed a dynamic model recommendation mechanism to ensure that each video frame is processed by the most suitable model. 
\begin{itemize}
    \item First, feature extraction is conducted to analyze video frames and extract key information of scene complexity, the number of objects, and the distribution of object sizes.
    \item Subsequently, we have established a compact model performance database to record the historical performance of each model across various feature scenarios.
    \item Finally, a recommendation algorithm is employed to recommend the optimal model for each video frame based on the extracted frame features and the information stored in the model performance database.
\end{itemize}

By implementing this model recommendation mechanism, the system is able to dynamically allocate tasks to the most suitable model for each video frame.

\subsection{2nd Team in MOSE Track: DeepSegMa}
  
\teaminfo{DeepSegMa}{Xuqiang Cao, Linnan Zhao, Jiaxuan Zhao, Fang Liu}{Key Laboratory of Intelligent Perception and Image Understanding, China}
  
\begin{figure}[t]
  \centering
  \includegraphics[width=\linewidth]{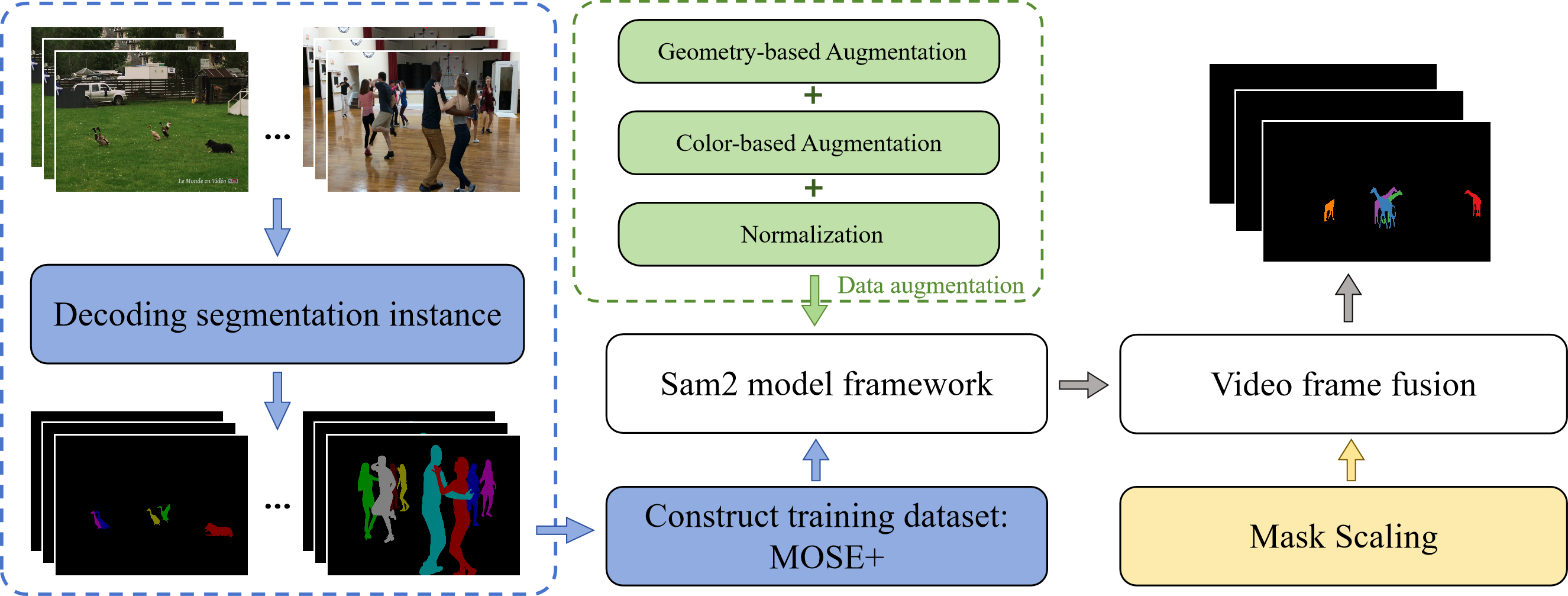}
  \caption{Overview of Team DeepSegMa's method.}
  \label{fig:m2_framework_all}
\end{figure}

\textbf{Method.}
An overview of our framework is presented in Figure~\ref{fig:m2_framework_all}. To better align with the characteristics of the MOSE dataset, we construct a tailored dataset, \textbf{MOSE+}, and introduce a set of targeted data augmentation strategies to mimic real-world variations in appearance, pose, illumination, and structural consistency. During inference, we employ a \textit{mask confidence control mechanism}, followed by temporal fusion across frames to generate the final segmentation outputs. Each component is detailed below.

\textbf{Baseline Model.}~We use a transformer-based segmentation framework with object-guided attention, mask-aware memory, and spatiotemporal reasoning. The model effectively captures temporal cues and spatial details through dual memory modules and multi-scale decoding, enabling robust performance under challenging scenarios like occlusion, motion blur, and small-object clutter.~This strong baseline lays a solid foundation for our enhancement strategies.

\textbf{Loss Function.}
To achieve high-precision segmentation and temporal consistency, we design a multi-task loss that combines pixel-wise accuracy, region-level overlap, classification discriminability, and robustness to occlusion. The total loss is defined as:
\begin{equation}
\mathcal{L}_{total} = \lambda_1 \mathcal{L}_{CE} + \lambda_2 \mathcal{L}_{Dice} + \lambda_3 \mathcal{L}_{Sim} + \lambda_4 \mathcal{L}_{MaskIoU},
\end{equation}
where $\mathcal{L}_{CE}$ denotes cross-entropy loss for foreground-background classification, $\mathcal{L}_{Dice}$ enhances region consistency, $\mathcal{L}_{Sim}$ enforces similarity between memory and query features, and $\mathcal{L}_{MaskIoU}$ constrains predicted mask quality. These losses are computed across multiple frames and candidate masks to jointly supervise spatiotemporal modeling.


\textbf{Data Augmentation.}
To improve generalization and robustness, we introduce a set of targeted augmentation strategies during training. Unlike static image tasks, video segmentation demands consistency across frames while simulating realistic variations. Our approach integrates both frame-consistent and frame-inconsistent perturbations:

\begin{itemize}
  \item \textbf{Consistent geometric transformations}: Random horizontal flipping, affine transformations (rotation, shear), and multi-scale resizing are applied across all frames in a clip to simulate viewpoint and object deformation.
  \item \textbf{Mixed color perturbations}: Brightness, contrast, and saturation changes are applied globally, while grayscale conversion and inconsistent color jittering are selectively applied to individual frames, enhancing robustness to lighting changes and visual ambiguity.
  \item \textbf{Normalization}: Images are transformed into tensors and normalized using ImageNet mean and standard deviation for stable convergence and pretrained compatibility.
\end{itemize}

These augmentations significantly improve the model's ability to handle structure variation, appearance change, and dynamic scenes in MOSE-like scenarios.

\textbf{Inference Strategy.}
To improve model robustness and adaptability in complex video scenarios, we introduce a set of tailored strategies during inference.

\textbf{Mask Confidence Control Strategy.} We observe that the quality of predicted masks can be significantly affected by post-processing in different scenarios, such as small objects, heavy occlusions, and target overlaps. To address this, we adopt a control strategy based on dynamic adjustment of the mask output distribution, using two key parameters: \textit{sigmoid scale} and \textit{sigmoid bias}. The sigmoid scale controls the sharpness of the output boundaries, while the sigmoid bias 
adjusts the overall activation level, thereby influencing the target coverage and boundary quality.
Experiments on the validation set show that setting the sigmoid scale to 7.5 and the sigmoid bias to -4.0 yields the best performance.

\textbf{Data.} To improve generalization and target modeling in complex scenarios, we construct an enhanced training set named \textbf{MOSE+}, based on the original MOSE dataset. This augmented set is composed of video segments from multiple public VOS datasets, selected to match the characteristics of MOSE, including frequent occlusions, dense small objects, object reappearance, and high similarity among targets. Specifically, we integrate carefully chosen sequences from datasets such as BURST~\cite{athar2023burst}, DAVIS~\cite{Pont-Tuset_arXiv_2017}, OVIS~\cite{qi2022occluded}, and YouTubeVIS~\cite{vos2019}, unify their annotations and resolution formats, and seamlessly merge them with MOSE to form a consistent training set, enhancing semantic understanding and robustness.

Please refer to the main technical report of DeepSegMa for model training and experiment details.

\subsection{3rd Team in MOSE Track: JIO}

\teaminfo{FVOS}{Mengjiao Wang, Junpei Zhang, Xu Liu, Yuting Yang, Mengru Ma}{International Joint Research Center for Intelligent Perception and Computation, China}

\label{sec:formatting}

\begin{figure}[!htbp]
	\centering
	\includegraphics[width=\linewidth]{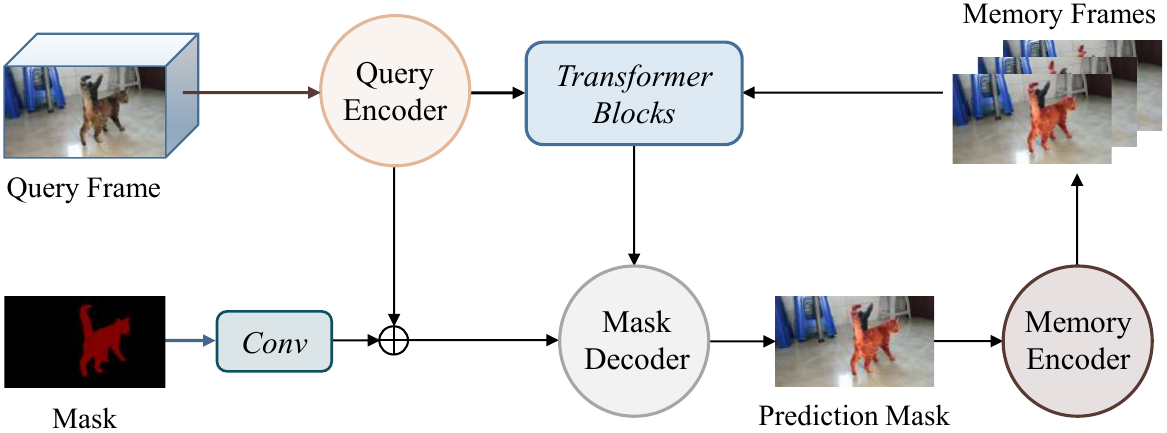}
	\caption{Network Architecture of FVOS.}
	\label{fig:m3_fig1}
\end{figure}

\begin{figure}[t]
	\centering
	\includegraphics[width=1\linewidth]{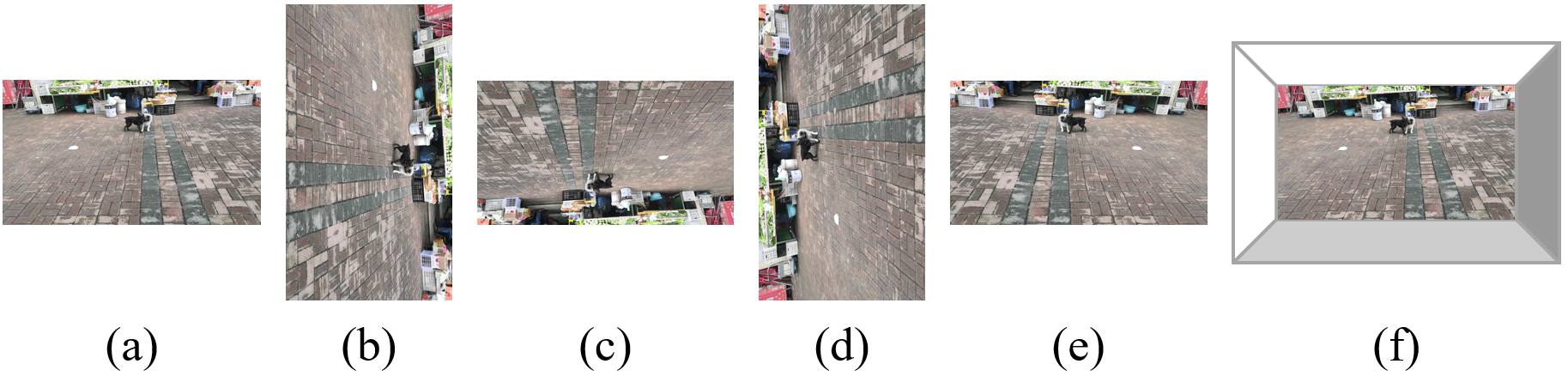}
    \vspace{-4mm}
	\caption{Test time data augmentation and multi-scale magnification operations. (a) original image. (b) clockwise by 90$^\circ$. (c) clockwise by 180$^\circ$. (d) clockwise by 270$^\circ$. (e) horizontal flipping. (f) multi-scale magnification.}
	
	\label{fig3}
\end{figure}

\textbf{Method.} Our approach primarily consists of three components: model fine-tuning training, morphological post-processing, and multi-scale segmentation result fusion.
Figure \ref{fig:m3_fig1} illustrates the network architecture adopted in our framework, which primarily relies on Transformers for feature extraction and attention computation. 

\textbf{MOSE Fine-tuning.}
\label{2.1}
Our training process is as follows: First, we fine-tune the pre-trained model on the MOSE dataset for a total of 10 epochs, submitting results from the validation set of each epoch. The best-performing model from this stage is selected as the pre-trained model to begin a new round of training. In this second stage, we conduct training for a total of 40 epochs, selecting the best-performing model for testing with optimal parameters. Finally, the single best-performing model is selected to generate the initial single-model segmentation results.

\textbf{Morphological Post-Processing.} After training,
we noticed that 
there exists a distinct gap between adjacent objects. This is because the model predicts separate objects individually before merging them during inference, thus the edge regions are not well aligned. To address this problem, we propose using morphological operations, especially dilation, for post-processing \cite{comer1999morphological}.

During the inference of the network, the binary segmentation masks for each object are first obtained and collected. For the current object, dilation operations are performed on both the object itself and all other objects. The adjacency between other objects and the current object is determined by checking whether the dilated masks overlap. If objects are deemed adjacent, the overlapping regions are filled and applied to the current object. Finally, object mask merging is performed following the rule of prioritizing higher-indexed objects, yielding the final segmentation results. 
Based on our experiments, using a kernel size of 2 yields better improvements in the segmentation results. 

\textbf{Multi-Scale Results Fusion. }
We also adopted common test-time data augmentation methods, including rotating the original image clockwise by 90$^\circ$, 180$^\circ$, and 270$^\circ$, horizontal flipping, as well as multi-scale processing by resizing the image to several scales, as shown in Figure \ref{fig3}. 
Specifically, starting from the original size, we resized the dataset with increments of 0.125 to reconstruct it at multiple scales. After experimenting with several scales, we finally selected 7 different scales ranging from 1 to 1.75 for fusion.

\section{MeViS Track Top Solution}
\label{sec:mevis_method}


\subsection{1st Team in MeViS Track: MVP-Lab}


\teaminfo{Unleashing the Potential of Large Multimodal Models for Referring Video Segmentation}{Hao Fang, Runmin Cong, Xiankai Lu, Zhiyang Chen, Wei Zhang}{Shandong University}

The input of RVOS contains a video sequence $\mathcal{S}$ = $\left\{X_t\in \mathbb{R}^{3 \times H \times W} \right\}_{t=1}^N $ with $N$ frames and a corresponding referring expression $\mathcal{T} = \left\{ T_l \right\}_{l=1}^L $ with \textit{L} words. 

\begin{figure}
  \centering
  \includegraphics[width=1.0\linewidth]{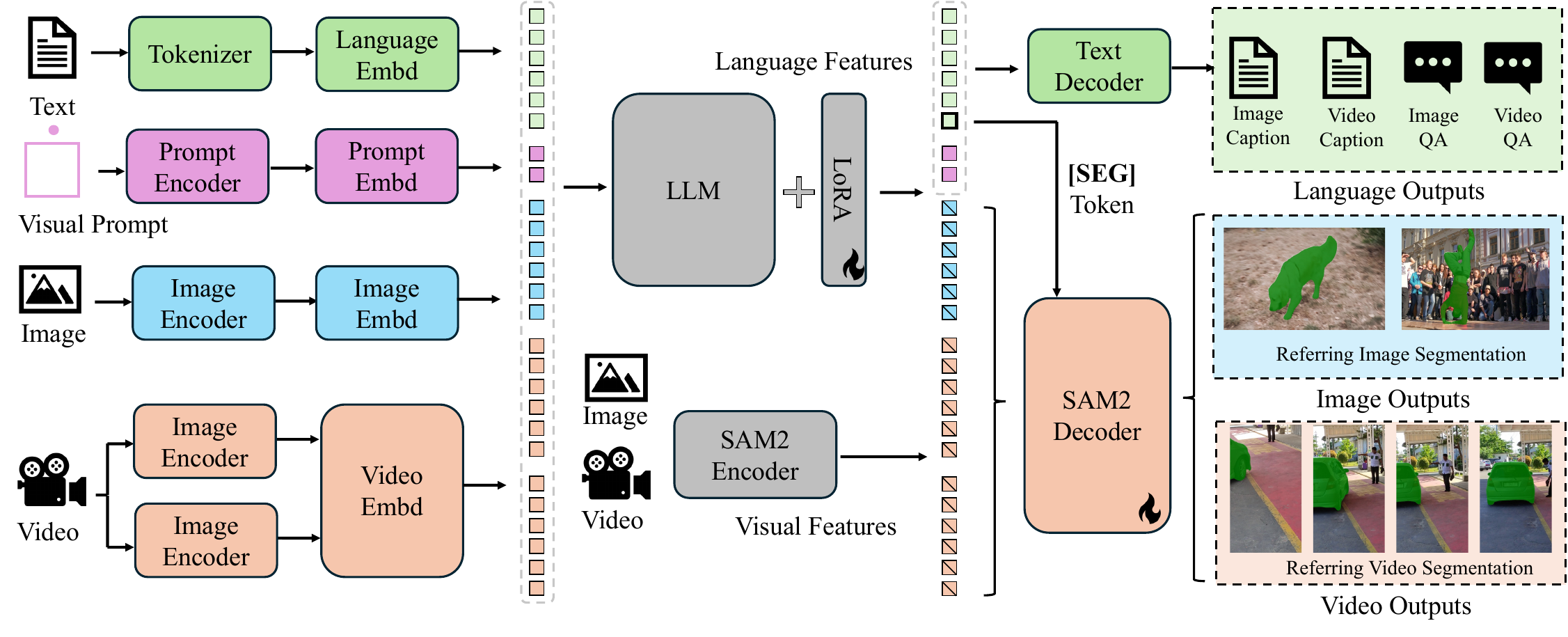}
  \vspace{-6mm}
  \caption{\textbf{The architecture of Sa2VA~\cite{yuan2025sa2va}.} The model first encodes the input texts, visual prompts, images, and videos into token embeddings. These tokens are then processed through a large language model (LLM). The output text tokens are used to generate the ``[SEG]'' token and associated language outputs. The SAM 2 decoder receives the image and video features from the SAM 2 encoder, along with the ``[SEG]'' token, to generate corresponding image and video masks.}
  \label{fig:r1_method}
\end{figure}

\textbf{Baseline.}
We adopt Sa2VA~\cite{yuan2025sa2va} as our baseline to obtain mask sequences $\mathcal{M} = \{M_t\}_{t=1}^N$ that are correlated with language descriptions:
\begin{equation}
    \mathcal{M} = \mathcal{F}^{rvos}\left( \mathcal{S}, \mathcal{T}\right),
\end{equation}
where $\mathcal{F}^{rvos}$ denotes the Sa2VA model. The overall architecture of Sa2VA is shown in Fig.~\ref{fig:r1_method}. It contains two parts: the LLaVA-like model and SAM 2. 

 \textbf{Pre-trained LMMs.} Sa2VA adopts pre-trained LLaVA-like models as the LMMs. It contains one visual encoder, one visual projection layer, and one LLM. The visual encoder takes input images, video, and sub-images as inputs. The visual projection layer maps inputs into visual tokens. These tokens, combined with the input text tokens, are the input of LLMs and the LLMs generate the text token prediction based on them. Note that Sa2VA adopts pre-trained LMMs following previous works~\cite{lai2024lisa,yan2024visa} to leverage their strong capability. It applies the same pipeline~\cite{wang2024qwen2} to both image and video chat datasets without modification.

 \textbf{Decoupled Design.} Sa2VA append SAM 2 alongside the pre-trained LLaVA model. It does not take the SAM 2's output tokens (visual features or decoder outputs) into LLM. There are three reasons.~1) Sa2VA makes the combination as simple as possible without increasing extra computation costs. 2) Adding extra tokens needs an extra alignment process. 3) Via this design, it can fully make our work as a plug-in-play framework to utilize pre-trained LMMs since the LMM community goes fast. Thus, Sa2VA adopts a decoupled design without introducing further communication between LLaVA and SAM 2.

 \textbf{Tuning SAM 2 Decoder via SEG Tokens.} Sa2VA connects SAM 2 and LMM via the special token ``[SEG]''. The hidden states of the ``[SEG]'' token are used as a new type of prompt and fed into SAM 2's Decoder to generate segmentation masks. The hidden states of ``[SEG]'' can be seen as a novel spatial-temporal prompt for SAM 2. SAM 2 segments the corresponding object mask in image and video based on the spatial-temporal prompt. During training, the SAM 2 decoder can be tuned to understand the spatial-temporal prompt, and gradients can be backpropagated through the ``[SEG]'' token to the LMM, allowing it to output the spatial-temporal prompt better.

\begin{algorithm}[!t]
\footnotesize
\caption{RVOS Inference Pipeline}\label{alg:refvos_inf}
\textbf{Input:} Video length $N$; Number of key frames $M$; Video frames $S_{N}$ ($X_1$, $X_2$, $X_3$,$\ldots$, $X_N$); Language description $T$;\\
\textbf{Output:} Sequence of masks $M_1$, $M_2$, $M_3$,$\ldots$, $M_N$;\\
\textbf{Run:} Sa2VA Model for RVOS;\\
Uniform sampling to extract key frames: $S_{M}$ $\gets$ $S_{N}$;\\
Visual embeddings: $E_v$ $\gets$ Encoder($S_{M}$);\\
Language embeddings: $E_l$ $\gets$ Encoder($T$);\\
Answers: $A$ $\gets$ LLM($\{E_v, E_l\}$);\\
Prompt embedding: $P_l$ $\gets$ Linear(Find($A$, '[SEG]'));\\
\For{$i = 1,2,\ldots,M$}{
    SAM 2 feature: $F_{i}$ $\gets$ Encoder($X_0$);\\
    Mask: $M_i$ $\gets$ Decoder($\{P_l, F_{i}\}$);\\
    Update Memory: $Mem$ $\gets$ Cross-Attention($\{Mem, M_i\}$);\\
}
\For{$i = M+1,M+2,\ldots,N$}{
    SAM 2 feature: $F_{i}$ $\gets$ Encoder($X_0$);\\
    Mask: $M_i$ $\gets$ Decoder($\{Mem, F_{i}\}$);\\
    Update Memory: $Mem$ $\gets$ Cross-Attention($\{Mem, M_i\}$);\\
}
\textbf{emit} $M_1$, $M_2$, $M_3$,$\ldots$, $M_N$;
\end{algorithm}

\textbf{Inference.}
\label{sec:inference} For RVOS tasks, Sa2VA designs a simple framework to achieve strong results on public benchmarks. In particular, for giving input video, it adopts a ``[SEG]'' token to generate the masks of the key frames. Then, it uses the memory encoded by the key frame features to generate the mask for the remaining frames. Sa2VA defaults to extracting the first five frames of the input video as key frames into LLM, but MeViS is a long video dataset, which results in a significant loss of video information. To address this, as shown in \cref{alg:refvos_inf}, we uniformly sample key frames across the entire video to provide the LLM with a more comprehensive temporal context.

These key frames are fed into CLIP and flattened to visual sequential tokens for LLM processing. The LLM takes the visual and language tokens as input and uses these tokens to extract information about the video to generate the ``[SEG]'' token. In SAM 2, the prompt encoder encodes boxes or clicks to prompt embeddings for object referring. Different from SAM 2, Sa2VA use two linear layers to project the ``[SEG]'' token into the language prompt embedding, which serves as an extension of the SAM 2 prompt encoders. With the language prompt embedding, it uses the SAM 2 decoder to generate the masks of the key frames. Then, Sa2VA use the memory encoder of SAM 2 to generate a memory based on the output key-frame masks. Finally, memory attention in SAM-2 uses this memory, along with prior non-key-frame masks, to generate the remaining frame masks.

\textbf{Aggregation.}
\label{sec:aggregation} We find that Sa2VA does not necessarily perform better with a larger number of parameters and more sampling frames, as each configuration has its own strengths in different videos. And for some videos that cannot be accurately segmented by LMMs, the classic RVOS model may handle them very well. So we integrate the results of multiple expert models to mitigate the erroneous predictions of a single model:
\begin{equation}
    \mathcal{M} = \mathcal{F}^{fuse}\left(\mathcal{M}^{K}\right),
\end{equation}
where $\mathcal{M}^{K}$ is the $K$ sets of mask sequences output by Sa2VA models with different configurations and other RVOS models~\cite{fang2024uninext}, $\mathcal{F}^{fuse}$ denotes pixel-level binary mask voting. If there are more than $(N+1)/2$ pixels with a value equal to 1, we divide the pixel into the foreground, otherwise, it is divided into the background.

\subsection{2nd Team in MeViS Track: ReferDINO-iSEE}


\teaminfo{ReferDINO-Plus: ReferDINO with SAM2}{Tianming Liang, Haichao Jiang, Wei-Shi Zheng, Jian-Fang Hu}{Sun Yat-sen University}

\begin{figure}[h]
    \centering
    \vspace{-6mm}
    \includegraphics[scale=0.16]{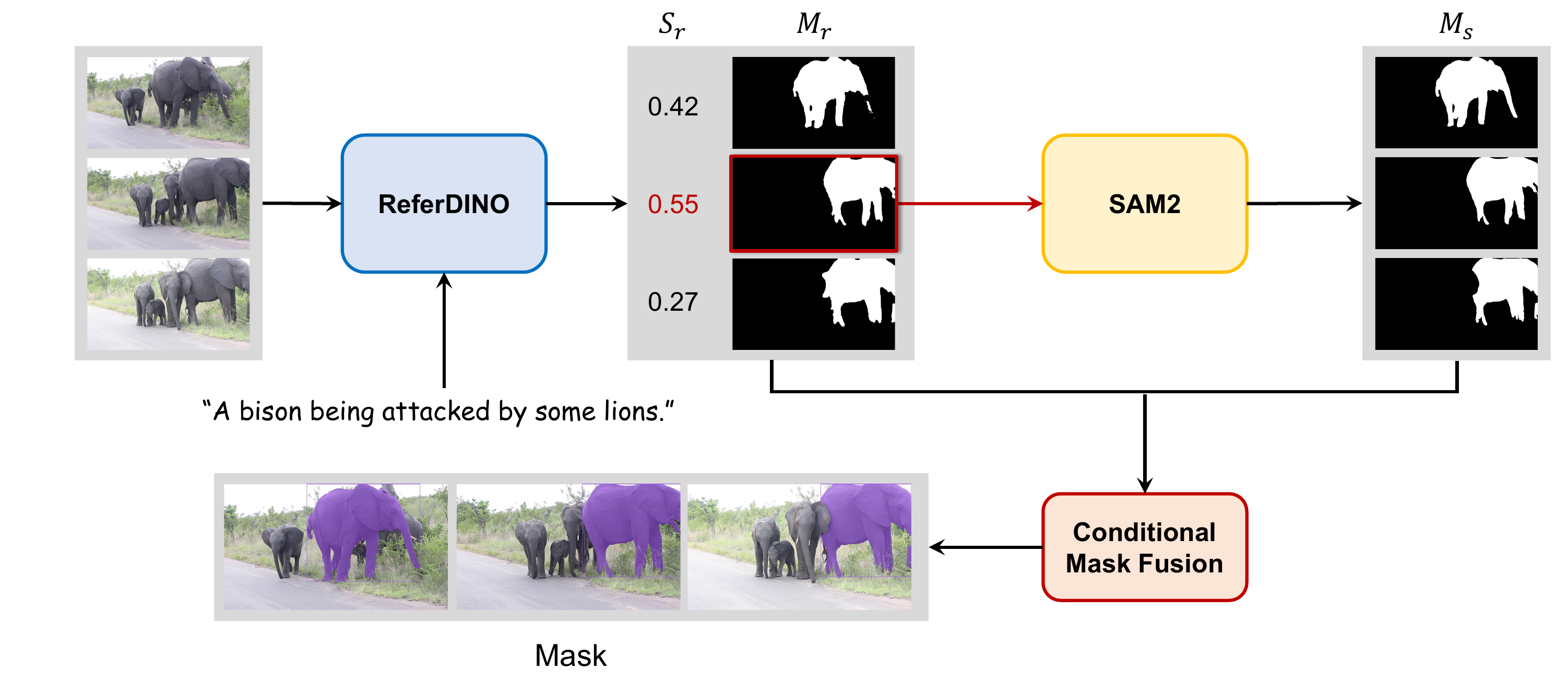}
    \caption{Overview of \textbf{ReferDINO-Plus}. For each video-description pair, we input it into ReferDINO to derive the object masks $M_r$ and the corresponding scores $S_r$ across the frames. Then, we select the mask with the highest score as the prompt for SAM2, producing refined masks $M_s$. Finally, we fuse the two series of masks through the \textit{conditional mask fusion} strategy.}
    \label{fig:r2_model}
 \end{figure}

The overall framework of our solution \textbf{ReferDINO-Plus} is presented in Figure~\ref{fig:r2_model}. For each video-description pair, we input it into ReferDINO to derive the object masks and the corresponding scores across the frames. Then, we select the mask with the highest score as the prompt for SAM2, producing refined masks. Finally, we fuse the two series of masks through the conditional mask fusion strategy, to generate the final masks for each frame.

\textbf{Cross-modal Dense Reasoning via ReferDINO.}
ReferDINO \cite{liang2025referdino} is a strong RVOS model inheriting object-level vision-language knowledge from GroundingDINO~\cite{liu2024grounding}, and is further endowed with pixel-level dense prediction and cross-modal spatiotemporal reasoning. Given a video clip of $T$ frames and a text description, ReferDINO performs cross-modal reasoning and segmentation, deriving a mask sequence $\{M_r^t\}_{t=1}^T$ and the corresponding scores $\{S_r^t\}_{t=1}^T$ throughout the video. Following previous works~\cite{liang2025referdino,MeViS,he2024decoupling}, we combine the multiple object masks with scores higher than a preset threshold $\sigma$ to handle multi-object cases.

\textbf{Post Enhancement with SAM2.}~SAM2~\cite{ravi2024sam} is a powerful prompt-based segmentation model capable of efficiently generating high-quality object masks across video frames given cues such as clicks, bounding boxes, or masks. We integrate SAM2 to enhance the mask precision and temporal consistency of ReferDINO predictions. After obtaining frame-wise masks and their associated confidence scores, we select the highest-scoring mask as a reference prompt. Using this reference frame and mask, SAM2 then propagates and refines the segmentation across the entire video, yielding a sequence of masks ${M_s^t}_{t=1}^T$.


\textbf{Conditional Mask Fusion.}
Although the masks from SAM2 are more reliable and stable, we observe that SAM2's overall performance on MeViS is significantly weaker than that of ReferDINO. In our experiments, we identify the main reason as that, for multi-object mask prompts, SAM2 tends to degenerate them into single-object masks, leading to substantial target loss in subsequent frames.
To address this issue, we design a \textit{Conditional Mask Fusion} (CMF) principle: for single-object cases, we output only the masks from SAM2; for multi-object cases, we combine both the masks from ReferDINO and SAM2. 

However, it remains challenging to determine whether an expression involves multiple objects. 
In our solution, we define it as a multi-object case if the mask area of SAM2 is less than $2/3$ of ReferDINO's. 
Formally, this process can be described as follows:
\begin{equation}
    M = 
    \begin{cases}
    M_s & \text{if} \  \mathcal{A}(M_s) < \frac{2}{3} \mathcal{A}(M_r), \\
    M_s + M_r & \text{otherwise},
    \end{cases}
\end{equation}
where $\mathcal{A}(\cdot)$ indicates the mask area. 
Note that our CMF is applied individually to each frame, which empirically achieves better performance.

\subsection{3rd Team in MeViS Track: Sa2VA}


\teaminfo{Sa2VA}{Haobo Yuan$^{1}$, Xiangtai Li$^{2}$, Tao Zhang$^{3}$, Lu Qi$^{2}$, Ming-Hsuan Yang$^{1}$}{{$^{1}$UC Merced}~{$^{2}$Bytedance}~{$^{3}$Wuhan University}}

  \textbf{Meta Architecture.} As shown in Fig.~\ref{fig:r1_method}, Sa2VA consists of an MLLM and SAM2. The MLLM accepts inputs of images, videos, and text instructions, and outputs text responses based on the text instructions. When the user instruction requires the model to output segmentation results, the text response will include the segmentation token ``[SEG]". The segmentation token's hidden states serve as implicit prompts and are converted through SAM2 into image and video-level object segmentation masks.
  
  \textbf{MLLM.} The SOTA MLLM InternVL 2.5~\cite{chen2024expanding} is adopted as the MLLM, demonstrating powerful capabilities in single-image, multi-image, and video understanding and conversation. 
  InternVL 2.5 adopts a LLaVA-like~\cite{liu2023visual} architecture, consisting of an InternVIT~\cite{chen2024internvl}, an MLP projector, and a Large Language Model. High-resolution images are first divided into several sub-images and a thumbnail, then encoded by InternVIT into vision tokens, which are mapped through one MLP and combined with text tokens as input to the LLM. The LLM will autoregressively output text responses, which may include segmentation tokens. The segmentation token's hidden states from the last LLM transformer layer are processed through an MLP to serve as the prompt input for SAM2~\cite{ravi2024sam}.
  \textbf{SAM2.} SAM2 generates object segmentation results for some high-resolution video frames based on the segmentation prompts output by the MLLM. Subsequently, SAM2 propagates these frame segmentation results to obtain object segmentation results for the entire video.
  
  \textbf{Sa2VA Model Training.} The original Sa2VA is co-trained on multiple datasets, including image/video VQA datasets, caption datasets, and image/video referring segmentation datasets, including MeViS. For this challenge, we do not fine-tune the model for MeViS, where we only focus on test time modifications on Sa2VA.

  \textbf{Naive Ref-VOS Inference Pipeline.} The origin pipeline of Sa2VA begins by extracting the first five frames ($k_1$, $k_2$, $\ldots$, $k_K$ are set to 1, 2, 3, 4, and 5 respectively) of the input video as keyframes, ensuring that they capture the critical context for the following processing. 
  These key frames are fed into CLIP and flattened to visual sequential tokens for LLM processing. 
  The LLM takes the visual and language tokens as input and uses these tokens to extract information about the video to generate the ``[SEG]'' token. 
  In SAM-2, the prompt encoder encodes boxes or clicks to prompt embeddings for object referring. 
  Different from SAM-2, we use two linear layers to project the ``[SEG]'' token into the language prompt embedding, which serves as an extension of the SAM-2 prompt encoders. 
  Using the language prompt embedding, we employ the SAM-2 decoder to generate key-frame masks.
We then encode these masks into memory via SAM-2’s memory encoder.
Finally, the memory attention module produces the remaining masks based on the key-frame and prior non-key-frame masks.
  %
  
  %
  
  \textbf{Test time augmentation for Sa2VA on MeVIS}
  \label{sec:method_test_time}
  \textbf{Long-Interleaved Inference.}
  The Naive Ref-VOS inference pipeline directly uses the first several frames as the keyframes. However, this may lead to suboptimal performance when the initial frames lack sufficient context for accurate reference embedding. This is especially evident when the language prompt requires a longer temporal reasoning.
  To address this issue, we propose an inference strategy named Long-Interleaved Inference (LII). We intentionally lengthen the time duration of the key frames to capture more context in the video. 
  Specifically, we interleave keyframes across a longer temporal window rather than selecting them consecutively from the beginning. 
  We sample keyframes at fixed intervals throughout the video, ensuring both early and late contextual signals are incorporated into the reference embedding.
  To keep the whole method simple and not overly dependent on hyperparameters, we use the same interval in all videos.
  The whole algorithm is similar to the naive Ref-VOS inference pipeline, and the main difference is the key frame selection strategy. $k_1$, $k_2$, $\ldots$, $k_K$ can be set to a fixed set of values before the execution of the entire pipeline.
  With the Long-Interleaved Inference strategy, the keyframes are no longer clustered at the beginning but are spread across a longer video clip. This design encourages the model to capture long-term dependencies, which is particularly beneficial in scenarios where the object appearance or scene context changes over time.
  
  \textbf{Other Attempts.} We also try a model ensembling strategy during the competition, which shows performance degradation and is not adopted in the final result. For the model ensembling strategy, we use two separate SAM-2 decoders during inference. The first one is from the Sa2VA, which is trained with the one-shot instruction tuning process and different from the original SAM-2 decoder as shown in Figure~\ref{fig:r1_method}.
  The other one is from the original SAM-2. 
  In the process of predicting the key frame masks, we have to use the SAM-2 decoder of Sa2VA because we need to use ``[SEG]'' token as prompt. We input the key frame masks into the second SAM-2 decoder to infer the rest of the masks.
  We hope to use this approach to separate reasoning and tracking. However, we observe a performance drop and do not apply this strategy.

\section{Conclusion and Discussion}
\label{sec:conclusion}

This year's PVUW challenge has attracted a record number of participants. This high level of engagement highlights the growing interest and relevance of pixel-level video understanding within the research community. From the top-performing methods, several key insights emerge.
First, we observe the critical importance of high-quality data. Datasets such as MOSE and MeViS, which offer fine-grained annotations, enable methods powered by large-scale pre-trained models like SAM 2 to achieve strong performance. Second, multi-modal large language models (LLMs) are beginning to demonstrate significant potential in video understanding, particularly in language-guided video tasks. With the continued evolution of LLMs, we expect them to play an increasingly vital role in this field.
These findings offer clear directions for future research. The importance of scaling—both in model capacity and the quality of training data—has been reinforced across many submissions. As LLMs continue to improve in multimodal capabilities, we believe they will further advance the state of video understanding.
Looking ahead, we will continue updating both the training and testing sets of the MOSE and MeViS datasets, and we remain committed to pushing the boundaries of pixel-level video understanding.

{
    \small
    \bibliographystyle{ieeenat_fullname}
    \bibliography{main}
}


\end{document}